\documentclass[letterpaper, 10 pt, conference]{ieeeconf}  

\usepackage[top=57pt,bottom=54pt,left=48pt,right=60pt]{geometry}

\IEEEoverridecommandlockouts                              

\usepackage{etoolbox} 
\usepackage[utf8]{inputenc} 
\usepackage[T1]{fontenc}    
\usepackage[english]{babel} 
\usepackage{comment}
\usepackage{cancel}
\usepackage{csquotes}
\usepackage{listings}
\usepackage{nameref}
\usepackage{paralist}
\usepackage{xspace}
\usepackage{setspace}
\usepackage{makeidx}
\usepackage{pdfpages}
\usepackage{upgreek}
\usepackage{flushend}

\usepackage{amsmath,amssymb} 
\usepackage{amsfonts}       
\usepackage{mathtools}
\usepackage{array} 
\usepackage{nicefrac}       
\usepackage{breqn}          
\usepackage{textcomp}
\usepackage{bm} 
\usepackage{pifont}
\usepackage[
  separate-uncertainty = true,
  multi-part-units = repeat
]{siunitx}

\usepackage{graphicx}
\usepackage{adjustbox} 
\usepackage{epsfig}

\usepackage{float}
\usepackage{subcaption}
\usepackage[font=small,labelfont=bf]{caption}
\usepackage{lscape}      

\usepackage{tikz}  
\usepackage{makecell}
\usepackage{overpic}
\usepackage{algorithm}
\usepackage[noend]{algpseudocode}

\usepackage{array} 
\usepackage{tabularx}
\usepackage{multirow}
\usepackage{multicol}
\usepackage{booktabs}   

\usepackage{paralist}
\let\labelindent\relax 
\usepackage{enumitem}
\setlist[itemize]{noitemsep,leftmargin=*,topsep=0in}
\setlist[enumerate]{noitemsep,leftmargin=*,topsep=0in}

\usepackage{url}            
\usepackage{color,colortbl} 
\usepackage{soul}

\PassOptionsToPackage{capitalize, noabbrev}{cleveref}
\usepackage{hyperref}
\hypersetup{pagebackref=false,breaklinks=true,colorlinks,urlcolor=blue,citecolor=blue,linkcolor=blue,bookmarks=false}

\makeatletter
\let\NAT@parse\undefined
\makeatother
\usepackage[numbers,sort&compress]{natbib}

\title{\LARGE \bf
GAGrasp: Geometric Algebra Diffusion for Dexterous Grasping
}

\author{Tao Zhong$^{1}$ and Christine Allen-Blanchette$^{1, 2}$
}

\begin{document}
\abovedisplayskip=4pt
\abovedisplayshortskip=4pt
\belowdisplayskip=4pt
\belowdisplayshortskip=4pt

\maketitle
\thispagestyle{empty}
\pagestyle{empty}

{
  \renewcommand{\thefootnote}%
    {\fnsymbol{footnote}}
  \footnotetext[0]{
$^{1}$Department of Mechanical and Aerospace Engineering, Princeton University, $^{2}$ Center for Statistics and Machine Learning, Princeton
University.
\\
Correspondence: {\tt\small \{tzhong, ca15\}@princeton.edu}}
}

\begin{abstract}
    We propose GAGrasp, a novel framework for dexterous grasp generation that leverages geometric algebra representations to enforce equivariance to $SE(3)$ transformations. By encoding the $SE(3)$ symmetry constraint directly into the architecture, our method improves data and parameter efficiency while enabling robust grasp generation across diverse object poses. Additionally, we incorporate a differentiable physics-informed refinement layer, which ensures that generated grasps are physically plausible and stable. Extensive experiments demonstrate the model's superior performance in generalization, stability, and adaptability compared to existing methods. Additional details at \href{https://gagrasp.github.io/}{gagrasp.github.io}
\end{abstract}

\section{Introduction}
\label{sec:intro}

Dexterous grasping remains a fundamental challenge in robotics, especially in unstructured environments where objects are encountered in diverse poses. Most existing datasets~\cite{turpin2023fast, goldfeder2009columbia,li2023gendexgrasp} contain grasps in canonical poses, which constrains learning-based grasp prediction methods to either predict grasps in the object frame~\cite{weng2024dexdiffuser,huang2023diffusion} or require non-trivial data augmentation~\cite{lundell2021ddgc, mayer2022ffhnet} to generate grasps in diverse environments. This limitation significantly hampers their applicability and scalability in real-world scenarios, where robots must interact with objects in arbitrary orientations, necessitating a robust approach to grasp generation that can generalize beyond the training data.

Recent advancements in diffusion models~\cite{weng2024dexdiffuser,huang2023diffusion} have shown promising results in generating high-quality dexterous grasps. However, these models often assume a canonical pose to reduce training effort, making them susceptible to out-of-distribution issues when applied to varied object orientations. Addressing this limitation requires an approach that inherently understands and leverages the symmetries present in the dexterous grasping problem.

Objects manipulated by robots typically undergo transformations described by the special Euclidean group $SE(3)$, (i.e. 3D rotations and translations). An ideal grasp generation model should exhibit equivariance to these transformations, meaning the probability of generating a successful grasp should remain invariant under any $SE(3)$ transformation of the object and the hand's base pose. Furthermore, the mapping from the object representation to the hand joint configurations should be invariant under $SE(3)$, ensuring consistent grasp quality regardless of object orientation.

In this context, geometric algebra emerges as a powerful and necessary tool. Given that a significant amount of data in robotics applications is geometric in nature, geometric algebra provides a unified framework~\cite{ruhe2023geometric,ruhe2024clifford} for handling such data. It offers a comprehensive language for describing geometric transformations and relationships, enabling the direct encoding of symmetry properties into the neural network architecture. By leveraging geometric algebra, we can achieve the desired equivariance and invariance more naturally and efficiently than with traditional methods.

In this work, we propose a novel framework that integrates geometric algebra to achieve the desired equivariance and invariance properties. Building on the principles outlined by~\citet{brehmer2024geometric}, we develop a symmetry-aware diffusion model for grasp generation that can effectively handle the diverse poses encountered in everyday manipulation tasks. This approach enhances our model's generalizability and improves data efficiency by embedding symmetry knowledge directly into the neural network architecture.
\begin{figure}[t]
  \centering
  \includegraphics[width=0.99\linewidth]{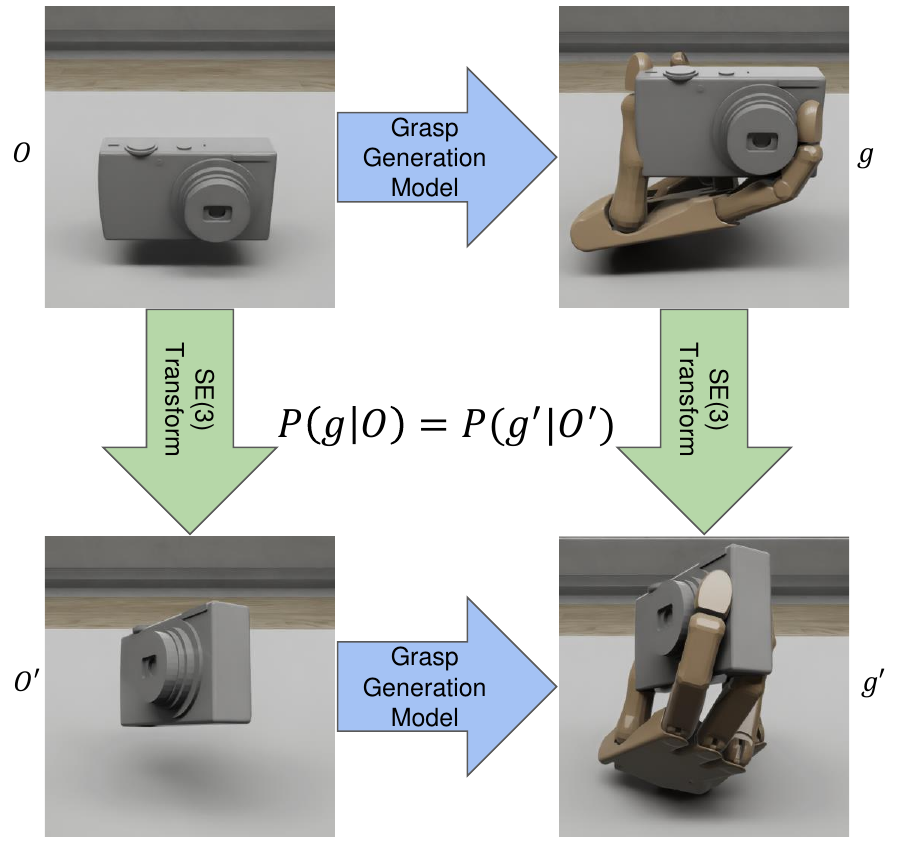}
  \captionsetup{belowskip=-20pt}
  \caption{\textbf{Symmetries in robotic grasping.} The figure shows how our model leverages $SE(3)$ symmetries. Starting with an object observation $O$, the model generates a grasp $\mathbf{g}$. After an $SE(3)$ transformation to $O'$, the probability of generating the corresponding transformed grasp $\mathbf{g}'$ remains invariant: $P(\mathbf{g}|O) = P(\mathbf{g}'|O')$.}
  \label{fig:teaser}
\end{figure}

To ensure that the generated grasps are not only geometrically feasible but also physically plausible, we incorporate a differentiable physics loss within a differentiable simulation engine. Existing vision-based grasp generation frameworks often require a refinement layer during inference to satisfy physical constraints. These methods often achieve this by incorporating an additional loss term to penalize physical violations~\cite{huang2023diffusion} or an adversarial loss term~\cite{weng2024dexdiffuser} to identify plausible grasps. Inspired by the classifier guidance used in diffusion models~\cite{dhariwal2021diffusion}, our differentiable physics loss function allows for the direct optimization of grasps. This approach ensures that the generated grasps adhere to physical constraints and are also stable and contact-rich.

Our contributions are threefold: (1) We introduce a symmetry-aware diffusion-based grasp generation framework that leverages geometric algebra for equivariance under the broader Euclidean $E(3)$ group. (2) We incorporate a differentiable physics loss to directly optimize the physical plausibility of generated grasps. (3) We demonstrate the effectiveness of GAGrasp through extensive experiments, showing improved generalization and data efficiency compared to existing methods.

\section{Related Work}
\label{sec:relatedwork}

\textbf{Vision-based Grasp Prediction} aims to learn a mapping from visual inputs to grasps by training on a dataset of positive examples. Traditional methods~\cite{varley2015generating, brahmbhatt2019contactgrasp,li2023gendexgrasp,lu2023ugg,wu2023learning} typically employ a two-stage approach by first generating a contact heatmap on the object's surface and then running an optimization algorithm~\cite{miller2004graspit} to fit the hand to the contact heatmap. Recent works have focused on end-to-end learning frameworks that can predict grasps directly using GANs~\cite{lundell2021ddgc,lundell2021multi,corona2020ganhand,mayer2022ffhnet}, VAEs~\cite{mousavian20196}, or direct regression~\cite{schmidt2018grasping,liu2020deep}. More recently, diffusion models have been explored for grasp generation~\cite{weng2024dexdiffuser,huang2023diffusion,urain2023se} due to their ability to generate high-quality samples from complex distributions. Despite these advancements, most existing methods~\cite{weng2024dexdiffuser,huang2023diffusion} assume a canonical pose for objects and grippers, which limits their applicability in real-world scenarios where objects can be encountered in arbitrary orientations. To address this, previous works have explored various data augmentation techniques~\cite{lundell2021ddgc,lundell2021multi,mayer2022ffhnet} to simulate diverse object poses during training. However, this approach always increases the computational burden and complexity of the training process. These limitations highlight the need for more robust approaches that can generalize to varied real-world conditions.

\textbf{Diffusion Models} have recently emerged as a powerful tool in the field of generative modeling, with applications spanning various domains, including robotics and grasping. In the context of robotics, diffusion models have been applied to tasks such as motion planning~\cite{carvalho2022conditioned,fang2023dimsam,carvalho2023motion,janner2022planning}, navigation~\cite{sridhar2023nomad}, and manipulation~\cite{chi2023diffusionpolicy,xian2023unifying,mishra2023reorientdiff}. Specifically for grasping,~\citet{urain2023se} propose a method for generating parallel-jaw grasp poses using diffusion models, demonstrating the effectiveness of diffusion processes in exploring the high-dimensional space of grasp configurations. Building on this foundation,~\citet{huang2023diffusion} and~\citet{weng2024dexdiffuser} introduce diffusion-based approaches for generating dexterous grasps. The use of diffusion models in grasp generation offers several advantages for their stable training dynamics and ability to produce high-quality samples. Moreover, their iterative nature allows for the incorporation of various constraints and guidance mechanisms during the generation process, making them well-suited for complex tasks like dexterous grasping. However, challenges remain in ensuring the physical plausibility of generated grasps and handling out-of-distribution poses. Addressing these challenges requires further research and the development of more sophisticated models and training techniques.

\textbf{Equivariant Neural Networks}
 learn representations that transform in a predictable way in response to specific transformations of the input. When these transformations are symmetries of the task, equivariant neural networks have been shown to both empirically~\cite{esteves2018learning,lei2023efem} and theoretically~\cite{elesedy2022group,zhu2021understanding} improve model generalization performance. Group equivariant convolutional neural networks~\cite{cohen2019general,kondor2018generalization} are a popular approach to equivariant neural network design. While this framework has been used to enforce equivariance to rotations~\cite{esteves2018learning,cohen2018spherical}, similarity transformations~\cite{lei2023efem,esteves2017polar}, and (special) Euclidean transformations~\cite{ryu2023diffusion,brehmer2024edgi}, it is restrictive in allowable input types and networks architectures~\cite{deng2021vector}. Recent works circumvent these limitations by representing network inputs and task symmetries using geometric algebra~\cite{ruhe2023geometric,brehmer2024geometric}. In our work, we use the geometric algebra framework to enforce the equivariance of grasp configurations to $SE(3)$ transformations of the input, yielding improved generalization performance in the presence of previously unseen input transformations.

\textbf{Differentiable Physics in Grasping} provides a promising way to ensure the physical plausibility of grasp configurations. Vision-based methods often incorporate heuristic-based refinement steps during inference to encourage contact~\cite{lundell2021ddgc,lundell2021multi} or enforce constraints such as non-penetration or force-closure~\cite{huang2023diffusion,weng2024dexdiffuser}. However, these approaches can be computationally expensive and may not guarantee optimal results. The integration of differentiable physics into grasp generation frameworks is well-explored in the area of differentiable grasp synthesis~\cite{liu2020deep,liu2021synthesizing,turpin2023fast,turpin2022grasp,li2023gendexgrasp}. Although these methods provide a more formal guarantee of physical plausibility, they are typically significantly slower and more computationally expensive than learning-based frameworks. In our work, we adapt and incorporate the differentiable metric proposed by Turpin et al.~\cite{turpin2023fast,turpin2022grasp} as a guidance signal during the generation process to allow for the direct optimization of grasp configurations. This approach ensures that the generated grasps are not only geometrically feasible but also adhere to physical constraints, resulting in stable and contact-rich grasps.

\section{Preliminaries}
\label{sec:prelim}

\subsection{Geometric Algebra}
Geometric algebra (GA) provides a unified framework for representing geometric objects and transformations. It extends traditional vector algebra by introducing multivectors, which can represent points, lines, planes, and higher-dimensional geometric entities.

A multivector in 3D GA $\mathbb{G}_{3,0,0}$ is expressed:
\begin{equation}
    x = x_s + \sum_{i=1}^3 x_i e_i + \sum_{i<j} x_{ij} e_i e_j + x_{123} e_1 e_2 e_3,
\end{equation}
where $x_s$, $x_i$, $x_{ij}$, and $x_{123}$ are real coefficients, and $e_i$ are basis vectors.  In the context of our work, we leverage the projective geometric algebra $\mathbb{G}_{3,0,1}$ following~\citet{ruhe2023geometric} and~\citet{brehmer2024geometric} which extends the 3D GA $\mathbb{G}_{3,0,0}$ with a fourth homogeneous coordinate $x_0 e_0$, resulting in a 16-dimensional multivector in the geometric algebra representation. The metric of $\mathbb{G}_{3,0,1}$ is defined such that $e_0^2=0$ and $e_1^2=e_2^2=e_3^2=1$, allowing for a more comprehensive representation of geometric objects and transformations~\cite{brehmer2024geometric}.

The fundamental operation in geometric algebra is the geometric product. For arbitrary multivectors $x,y,z \in \mathbb{G}_{3,0,1}$ and scalar $\lambda$, the geometric product satisfies: (i) closure, i.e., $xy\in \mathbb{G}_{3,0,1}$, (ii) associativity, i.e., $(xy)z=x(yz)$, (iii) commutative scalar multiplication, i.e., $\lambda x = x\lambda$, (iv) distributivity over addition, i.e., $x(y+z)=xy+xz$, and (v) vectors square to scalars given by a metric norm. It can also be deduced that the geometric product of two basis vectors is antisymmetric, i.e., $e_i e_j=-e_j e_i$ for $i\neq j$.

The projective geometric algebra $\mathbb{G}_{3,0,1}$ allows us to represent geometric objects and transformations uniformly, as described in Table~\ref{tab:ga_embed}. For a transformation $u$ and a geometric object $v$ both represented as multivectors in $\mathbb{G}_{3,0,1}$, the action of $u$ on $v$ is given by the sandwich product $u[v] = uvu^{-1}$.

For a more comprehensive study of geometric algebra and its applications, we direct readers to~\cite{dorst2007geometric, brehmer2024geometric, ruhe2023geometric, ruhe2024clifford}.

\subsection{Problem Formulation}
\begin{table}[t]
\centering
\small
\resizebox{\columnwidth}{!}{\begin{tabular}{lll}
\toprule
\textbf{Transformation} & \textbf{Geometric Objects} & \textbf{$\mathbb{G}_{3,0,1}$ Element} \\ \midrule
Identity & Scalar & Scalar \\
Reflection & Plane & Vector \\
Rotation / Translation & Line & Bivector \\
Roto / Transflection & Point & Trivector \\
Screw & Pseudoscalar & Quadvector \\
\bottomrule
\end{tabular}}
\captionsetup{belowskip=-20pt}
\caption{Embeddings of group elements in $E(3)$ and common geometric objects in $\mathbb{G}_{3,0,1}$. Each transformation or geometric object can be represented by components of the multivector.}
\label{tab:ga_embed}
\end{table}
We consider the problem of generating dexterous grasps for objects represented by point clouds. Given a full point cloud observation $O \in \mathbb{R}^{N\times 3}$, our goal is to sample dexterous grasps $G$ that parameterize a dexterous hand stably grasping the object. Each grasp $\mathbf{g} \in G \in \mathbb{R}^{9+k}$ is represented by a tuple $[\mathbf{r}, \mathbf{p}, \mathbf{q}]$, where $\mathbf{r} \in \mathbb{R}^6$, $\mathbf{p} \in \mathbb{R}^3$ are the rotation (as in~\cite{zhou2019continuity}) and translation of the hand base, respectively, and $\mathbf{q} \in \mathbb{R}^k$ denotes the joint configurations of the hand. The grasp generation problem can be framed as learning a distribution $p(\mathbf{g} \mid O)=p([\mathbf{r}, \mathbf{p}, \mathbf{q}] \mid O)$ over grasps conditioned on the observed point cloud. Our objective is to develop a model that can efficiently sample high-quality grasps from $p(\mathbf{g} \mid O)$.

\subsection{Equivariance and Invariance in Grasp Generation}
In the context of grasp generation, $SE(3)$ equivariance is a desirable property. A function $f:X\rightarrow Y$ is said to be equivariant with respect to the transformation group $\mathcal{G}$ if $f(\rho_X(x)) = \rho_Y(f(x))$ for a group element $\rho$ and an input $x$, where $\rho_Z$ denotes the action of the group element $\rho$ on the space $Z$. Invariance can be described as a special case of equivariance where the group action $\rho_Y$ is the identity transformation and can be mathematically defined as $f(\rho_X(x)) = \rho_Y(f(x)) = f(x).$

In our grasp generation framework, we require that the mapping from the object representation to the hand base pose be equivariant to $SE(3)$ transformations and that the mapping from the object representation to the hand joint configurations be invariant to $SE(3)$ transformations. Probabilistically, this can be written as:
\begin{equation}
    p([\mathbf{r},\mathbf{p},\mathbf{q}] \mid O) = p([\rho \cdot (\mathbf{r},\mathbf{p}),\mathbf{q}] \mid \rho_{\mathbb{R}^3}(O)),
    \label{eq:symm}
\end{equation}
where $\rho\in SE(3)$, and $g \cdot h$ denotes the composition of group elements $g,\text{and }h\in\mathcal{G}$.


\section{Methodology}
\label{sec:method}
\begin{figure*}[ht!]
  \centering
  \includegraphics[width=0.99\linewidth]{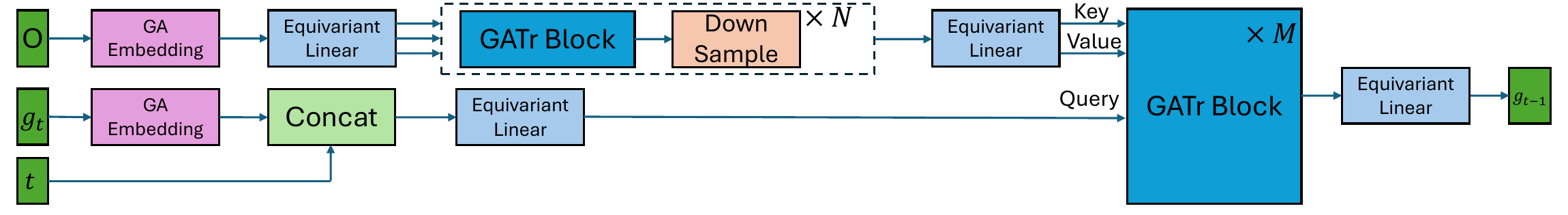}
  \captionsetup{belowskip=-11pt}
  \caption{\textbf{Overview of the GAGrasp architecture.} The point cloud $O$ and grasp configuration $g_t$ are embedded using $\mathbb{G}_{3,0,1}$, processed through GATr blocks ensuring $SE(3)$ equivariance, with down-sampling to enhance efficiency. A cross-attention mechanism uses the embedded grasp and diffusion step $t$ to predict the updated grasp $g_{t-1}$.}
  \label{fig:architecture}
\end{figure*}
In this section, we introduce our framework for generating dexterous grasps. The framework is composed of a grasp sampler that leverages a conditional diffusion model, which is designed to handle the high-dimensional space of dexterous grasps while ensuring robustness to variations in object poses.

\subsection{Diffusion-based Grasp Generation}
Our diffusion-based grasp generation pipeline is designed to generate high-quality dexterous grasps conditioned on the observed point cloud of the object. The grasp generator operates by iteratively refining randomly sampled grasps through a diffusion process, ensuring the final output is feasible and optimized for grasp success.

The grasp generator leverages a conditional diffusion model~\cite{ho2020denoising}. The process begins with the forward diffusion of a successful grasp $\mathbf{g}_0 = [\mathbf{r}_0, \mathbf{p}_0, \mathbf{q}_0]$, which includes the 6D rotation vector $\mathbf{r}_0$, the 3D position $\mathbf{p}_0$, and the hand joint configurations $\mathbf{q}_0$, into Gaussian noise over $T$ timesteps. Formally, the forward diffusion process is defined:
\begin{equation}
   q(\mathbf{g}_{1:T} \mid \mathbf{g}_0) = \prod_{t=1}^T \mathcal{N}(\mathbf{g}_t; \sqrt{1-\beta_t} \;\mathbf{g}_{t-1}, \beta_t I), 
\end{equation}
where $\beta_t$ is the scheduled noise variance at timestep $t$.

To recover the original grasp $\mathbf{g}_0$ from the noisy grasp $\mathbf{g}_T$, the model iteratively estimates and removes the added Gaussian noise using a noise predictor $\epsilon_\theta$, parameterized by $\theta$. The training objective for the noise predictor is to minimize the loss:
\begin{equation}
    L_{\epsilon} = \left\lVert \epsilon_\theta(\mathbf{g}_t, O, t) - \epsilon_t \right\rVert^2_2,
\end{equation}
where $\epsilon_t$ is the ground-truth noise at timestep $t$.

To sample from $p_\theta(\mathbf{g}_0 \mid O)$, the generator performs an iterative denoising process starting from the noisy grasp $\mathbf{g}_T$ and the point cloud $O$, which can be defined as:
\begin{equation}
    p_\theta(\mathbf{g}_{0:T} \mid O) = p(\mathbf{g}_T) \prod_{t=1}^T \mathcal{N}(\mathbf{g}_{t-1}; \boldsymbol{\mu}_t, \boldsymbol{\Sigma}_t),
    \label{eq:backward}
\end{equation}
where $\boldsymbol{\mu}_t = \mu_\theta(\mathbf{g}_t, O, t)$ and $\boldsymbol{\Sigma}_t = \Sigma_\theta(\mathbf{g}_t, O, t)$ can be inferred from $\epsilon_\theta(\mathbf{g}_t, O, t)$ and the parameters of the diffusion process.

\subsection{Equivariant Model Architecture}
\begin{figure*}[ht!]
  \centering
  \begin{minipage}{0.72\linewidth}
      \includegraphics[width=0.99\linewidth]{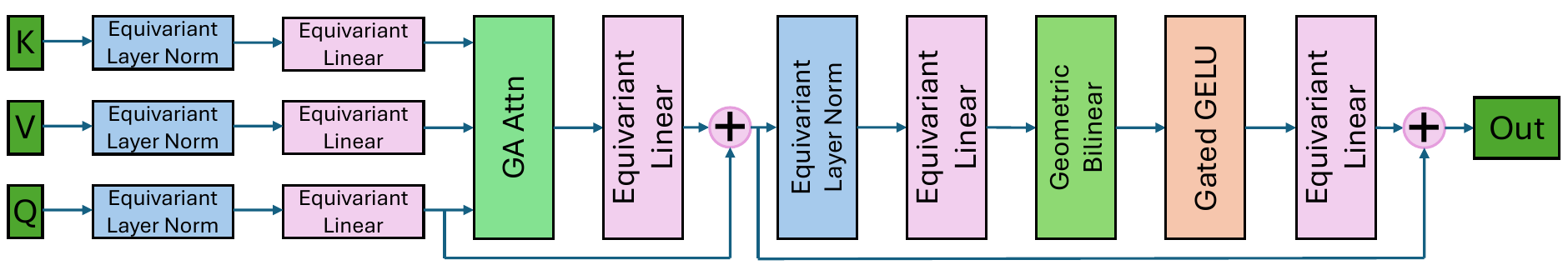}
  \end{minipage}
  \begin{minipage}{0.27\linewidth}
  \caption{\textbf{The GATr Block} processes key, value, and query inputs with equivariant layers, using geometric algebra-based attention and nonlinearities to maintain $SE(3)$ equivariance of the grasp configuration.}
  \label{fig:gatr}
  \end{minipage}
  \vspace{-18pt}
\end{figure*}
As shown in~\cite{kohler2020equivariant,bose2021equivariant,papamakarios2021normalizing,hoogeboom2022equivariant}, if $p(\mathbf{g}_T)$ is invariant and the neural network denoising model $p_\theta(\mathbf{g}_{t-1} \mid \mathbf{g}_{t}, O)$ is equivariant, then the marginal distribution $p_\theta(\mathbf{g}_{0} \mid O)$ of the denoising model will be an invariant distribution, satisfying the desired symmetries in Eq.~\ref{eq:symm}. Building on this principle, we leverage the primitives defined by~\citet{brehmer2024geometric} acting on multivectors to incorporate symmetries into the model architecture, ensuring that the grasp generation process respects the geometric transformations of the input data. Figure~\ref{fig:architecture} illustrates our model architecture. At each iteration, we first embed all data into multivector representation in $\mathbb{G}_{3,0,1}$. The multivectors are then processed through equivariant layers operating on multivector inputs, as shown in Figure~\ref{fig:gatr}. Finally, we extract the desired geometric quantities from the geometric features for loss calculation and backpropagation. Below, we detail the construction of the equivariant modules.

\textbf{Equivariant Linear Layers} are fundamental components that ensure the geometric properties of the input data are preserved through transformations.~\citet{brehmer2024geometric} show that a linear map $\phi: \mathbb{G}_{3,0,1} \rightarrow \mathbb{G}_{3,0,1}$ is equivariant if it is of the form:
\begin{equation}
    \phi(x) = \sum_{k=0}^{4} w_k \langle x \rangle_k + \sum_{k=0}^3 v_k e_0 \langle x \rangle_k ,
\end{equation}
where $\langle x \rangle_k$ is the grade projection that isolates the $k$-grade components of the multivector $x$, and $w_k$ and $v_k$ are learnable parameters.

\textbf{Geometric Bilinears Layers} handle interactions between multivectors, which combine the geometric product and the join operation. The geometric product $xy$ allows for the mixing of different grades, and the join operation $\text{Join}(x,y) = (x^* \wedge y^*)^*$, where $x^*$ is the dual of $x$ and $x\wedge y$ denotes the exterior (wedge) product between $x$ and $y$, is necessary for constructing meaningful geometric features. Combining the geometric product and the join, the geometric bilinear layer is defined:
\begin{equation}
    \text{Geometric}(x,y; z) = \text{Concat}(xy, z_{0123} (x^* \wedge y^*)^*),
\end{equation}
where $z_{0123}$ is the pseudoscalar component of a reference multivector $z$ calculated from the network inputs.

\textbf{Equivariant Attention Layers} extend the standard dot-product attention to operate on multivectors while respecting the $SE(3)$ symmetry. Given multivector-valued query, key, and value tensors $(Q,K,V)$ with $n_i$ tokens and $n_c$ channels, the attention mechanism computes the attention scores using the inner product in $\mathbb{G}_{3,0,1}$:

\footnotesize
\begin{equation}
    \text{Attention}(Q, K,V)_{i' c'} = \sum_i \text{Softmax} \left ( \frac{\sum_c \langle Q_{i' c'}, K_{i c'} \rangle}{\sqrt{8 n_c}}\right ) V_{ic'} ,
\end{equation}
\normalsize
where $i, i'$ index the tokens, $c, c'$ index the channels, and $\langle \cdot, \cdot \rangle$ denotes the invariant inner product of the multivectors, which is the regular dot product on the 8 of the 16 dimensions that do not contain $e_0$. Cross-attention and multi-head attention are calculated using this mechanism in the usual way~\cite{vaswani2017attention}.

\textbf{Equivariant Nonlinearities and Normalization}
For nonlinear transformations, we follow~\citet{brehmer2024geometric} and use scalar-gated GELU (Gaussian Error Linear Unit) nonlinearities~\cite{hendrycks2016gaussian}, defined as: $\text{GatedGELU} (x) = \text{GELU} (x_1) x$, where $x_1$ is the scalar component of the multivector $x$.

Normalization is achieved through an E(3)-equivariant LayerNorm, defined as:
\begin{equation}
    \text{LayerNorm} (x) = \frac{x}{\sqrt{\mathbb{E}_c [\langle x, x \rangle]}}
\end{equation}
where the expectation $\mathbb{E}_c$ of the inner product is taken over the channels.

\textbf{Down-Sampling Layers} are inspired by the Transition Down module from PointTransformer~\cite{zhao2021point}. While the original GATr framework~\cite{ruhe2023geometric} maintains identical input and output sizes, this approach can lead to computational inefficiency, particularly in large-scale applications. To address this, we integrate a down-sampling mechanism that reduces the cardinality of the point set, improving computational efficiency.

The Down-sampling module begins with farthest point sampling (FPS) in the $xyz$ space to select a well-distributed subset of points. We then construct a k-nearest neighbors (kNN) graph to pool features from the original point set onto the down-sampled subset. The max pooling operation is performed on the scalar component of the multivectors, selecting the multivector with the largest scalar component, which is then propagated forward.

\textbf{Symmetry Breaking}
All the modules described above are equivariant to $E(3)$ transformations of the hand and object and permutations $S_n$ of the object point cloud. However, full $E(3)$ symmetry may introduce unnecessary inductive biases. In grasping problems, we are more interested in $SE(3)$ symmetries because of the hand morphology. To resolve unnecessary inductive biases, we include symmetry-breaking mechanisms. For example, we break the chirality symmetry by introducing pseudoscalar features to encode handedness, enabling the model to better adapt to specific tasks. 

\subsection{Differentiable Physics-Informed  Refinement Layer}
To ensure the physical plausibility and stability of the generated grasps, we integrate a differentiable physics-informed refinement layer into our framework. Inspired by Turpin et al.~\cite{turpin2023fast,turpin2022grasp}, the refinement layer uses gradients obtained from a differentiable physics simulation engine~\cite{warp2022,heiden2021disect} to iteratively adjust the grasp configuration during the denoising process.

Given a grasp $\mathbf{g} = [\mathbf{r}, \mathbf{p}, \mathbf{q}]$, the optimization process aims to minimize a physics-informed loss function $L_{\text{phys}}$ calculated in a differentiable physics simulator, which encapsulates the aforementioned constraints. Specifically, we set an initial object velocity $\dot{\textbf{p}}_{\text{obj}}(0)$ and test whether contact with the static grasp set to $\mathbf{g}$ can dampen it. The object is initialized in the pose $(\textbf{r}_{\text{obj}}(0), \textbf{p}_{\text{obj}}(0))$ from the dataset. We simulate $T_{\text{sim}}$ timesteps in a physics engine and compute the object's final translational and angular velocities $(\dot{\textbf{r}}_{\text{obj}}(T_{\text{sim}}), \dot{\textbf{p}}_{\text{obj}}(T_{\text{sim}}))$. To comprehensively evaluate stability, we perform this test across a batch of simulations with different initial velocities $\dot{\textbf{p}}_{\text{obj}}(0)_m$. The stability loss we are trying to optimize is then given by:

\vspace{-5pt}
\small
\begin{equation}
    L_{\text{stability}} = \frac{1}{M} \sum_{m=1}^M \left ( \left\lVert \dot{\textbf{p}}_{\text{obj}}(T_{\text{sim}})_m \right\rVert^2_2 + \left\lVert \dot{\textbf{r}}_{\text{obj}}(T_{\text{sim}})_m \right\rVert^2_2 \right )
\end{equation}
\normalsize
where $M$ is the total number of simulations, each with a different initial velocity.

Besides the stability test, we also include two additional loss terms to encourage the joint configurations to stay within the joint limits:

\vspace{-5pt}
\footnotesize
\begin{equation}
    L_{\text{range}} = \left\lVert \textbf{q} - \frac{\textbf{q}^{\text{up}} + \textbf{q}^{\text{low}}}{2} \right\rVert^2_2, \ L_{\text{limit}} = \max (\textbf{q} - \textbf{q}^{\text{up}}, 0) + \max (\textbf{q} - \textbf{q}^{\text{low}}, 0)
\end{equation}
\normalsize
where $\textbf{q}^{\text{up}}$ and $\textbf{q}^{\text{low}}$ denote the upper and lower limits of the joints, respectively.

The overall objective function is $L_{\text{phys}} = L_{\text{stability}} + \alpha_1 L_{\text{range}} + \alpha_2 L_{\text{limit}}$, where $\alpha_1=0.01$ and $\alpha_2=10$ are hyperparameters. We follow the simulation and optimization techniques outlined in~\citet{turpin2023fast} to address the issues of contact sparsity and local flatness of triangular meshes.

We then modify the denoising process to incorporate the gradient-based adjustments at each timestep. Specifically, the process iteratively refines the grasp configuration using $\hat{\boldsymbol{\mu}}_t = \boldsymbol{\mu}_t + \lambda \nabla_g L_{\text{phys}}$ where $\boldsymbol{\mu}_t$ is defined as in Equation~\ref{eq:backward}, and $\lambda$ is a scaling factor that controls the influence of the physics-based optimization.

\section{Experiments}
\label{sec:exp}
\begin{figure*}[ht!]
  \centering
  \begin{minipage}{0.7\linewidth}
  \includegraphics[width=0.99\linewidth]{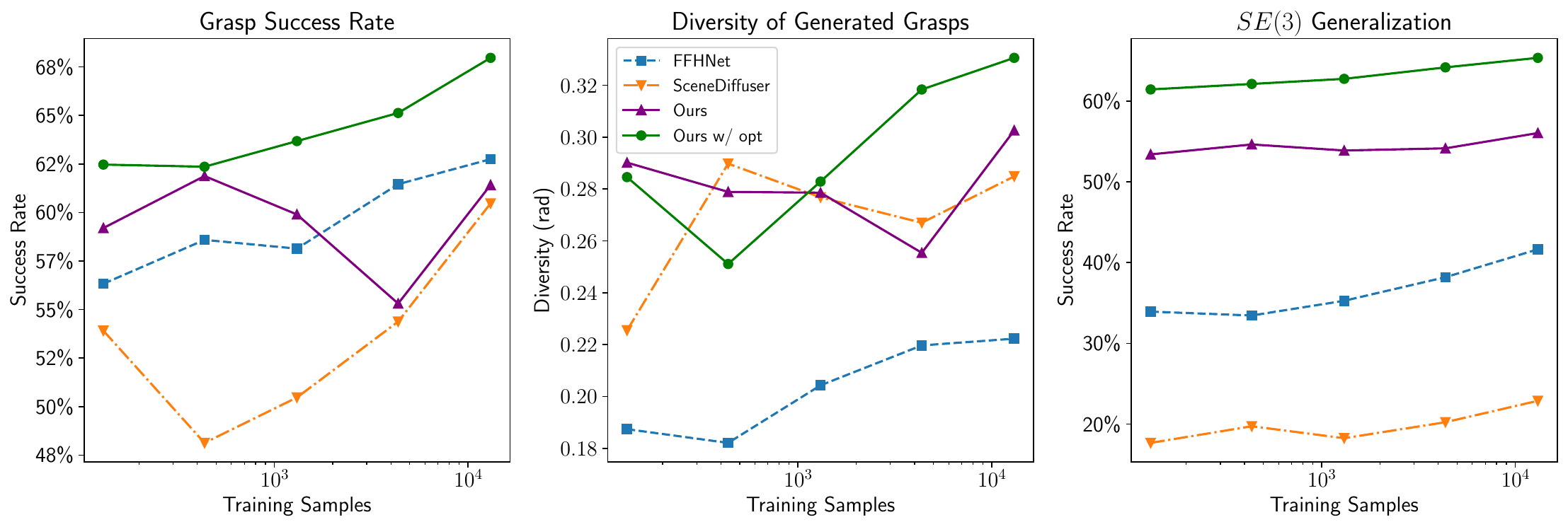}
  \end{minipage}
  \begin{minipage}{0.29\linewidth}
  \caption{\textbf{Experimental Results.} Grasp success rate and diversity metrics across different data amounts. "Ours w/ opt" refers to our model with the physics-informed refinement layer. \textbf{Left:} Our model outperforms others in grasp success rate, especially with fewer training samples. \textbf{Middle:} Our model generates more diverse grasps. \textbf{Right:} Our method is robust to out-of-distribution data with random $SE(3)$ transformations.}
  \label{fig:results}
  \end{minipage}
  \vspace{-12pt}
\end{figure*}

\begin{figure*}[ht!]
  \centering
  \begin{minipage}{0.78\linewidth}
  \includegraphics[width=0.99\linewidth]{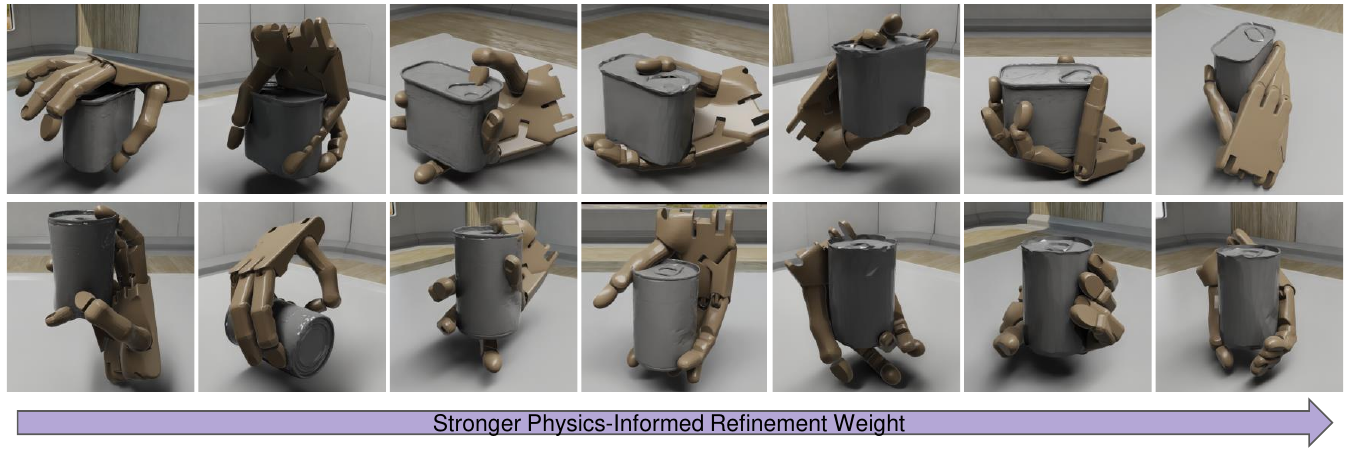}
  \end{minipage}
  \begin{minipage}{0.21\linewidth}
  \caption{\textbf{Example Grasps Generated by our Method.} Our model generates stable grasps for unseen objects. As the physics-informed refinement weight $\lambda$ increases (left to right), the model produces more stable power grasps, while a smaller $\lambda$ results in precision fingertip grasps, showing adaptability in grasp types.}
  \label{fig:qual}
  \end{minipage}
  \vspace{-15pt}
\end{figure*}
In this section, we investigate the effectiveness of GAGrasp through a series of experiments designed to address the following research questions: (1) Does the use of GA representation enhance data and parameter efficiency? (2) How does GAGrasp perform on unseen data with random $SE(3)$ transformations applied to test data but not training data? (3) How much does the physics-informed refinement layer contribute to grasp quality and stability?

\subsection{Experimental Setup}
\textbf{Dataset} We follow~\citet{huang2023diffusion} to use the Shadowhand subset of the MultiDex~\cite{li2023gendexgrasp} dataset, which includes 16,069 dexterous grasping poses for 58 daily objects. The dataset is split into 48 training objects and 10 unseen test objects. The grasp configuration of the Shadowhand is represented by $\mathbf{g} = [\mathbf{r}, \mathbf{p}, \mathbf{q}] \in \mathbb{R}^{33}$, where $\mathbf{r} \in \mathbb{R}^6$, $\mathbf{p} \in \mathbb{R}^3$ are the rotation and translation of the hand base, respectively, and $\mathbf{q} \in \mathbb{R}^{24}$ denotes the joint configurations of the hand. Objects are represented by their point cloud $O \in \mathbb{R}^{2048 \times 3}$ sampled with 2,048 points.

\textbf{Metrics} We evaluate models based on the success rate of the sampled grasps in IssacGym~\cite{liang2018gpu} following the setup from~\citet{li2023gendexgrasp}. Specifically, we test if a grasp is successful in the simulation by applying external forces in 6 directions along the $\pm xyz$ axis to the object and measuring its movement. A grasp is considered successful if it withstands all six tests without moving more than $2\text{cm}$ after applying a consistent $0.5\text{m/s}^2$ acceleration over 60 simulation steps. We use a friction coefficient of 10, an object density of 10,000, and IsaacGym’s built-in positional controller to achieve the target joint configuration. We also report the diversity score as the mean standard deviation among all revolute joints.

\textbf{Baseline Models}
We compare our method with a diffusion-based model SceneDiffuser~\cite{huang2023diffusion}, a cVAE-based method FFHNet~\cite{mayer2022ffhnet}, and a version of our model without the physics-informed refinement layer.

\subsection{Geometric Algebra Representation Efficiency}
To assess data and parameter efficiency conferred by the GA representation, we train our model and baseline models with varying amounts of training data, measuring grasp success rate across different data regimes. As shown in the left plot of Figure~\ref{fig:results}, our model consistently outperforms the baselines, particularly in low-data regimes, where data efficiency is critical. Even with limited training samples, our method achieves a higher success rate, indicating that the GA representation enables more effective learning. Our method also generates more diverse grasps due to the use of a diffusion-based approach, as evidenced by the diversity scores in the middle plot. Despite having approximately 20\% of the learnable parameters of FFHNet~\cite{mayer2022ffhnet}, our model, without the physics-informed refinement layer, performs on par with it and consistently outperforms SceneDiffuser~\cite{huang2023diffusion} across all training sample sizes. Notably, the slight performance drop observed with larger datasets can be attributed to the generative nature of our method, which keeps success rates robust while increasing grasp diversity, and to the variance in the physical simulation-based evaluation. 

\subsection{Performance on Data with Out-of-Distribution Poses}
To evaluate the model's robustness to data with unseen poses, we apply random $SE(3)$ transformations to the test set objects, simulating scenarios where objects are encountered in poses not seen during training. As shown in the right plot of Figure~\ref{fig:results}, our method significantly outperforms both SceneDiffuser~\cite{huang2023diffusion} and FFHNet~\cite{mayer2022ffhnet} on this challenging task. SceneDiffuser~\cite{huang2023diffusion} performs the worst because it is trained with objects in a canonical pose, making it less adaptable to new orientations. FFHNet~\cite{mayer2022ffhnet} performs better than SceneDiffuser~\cite{huang2023diffusion} but still falls short of our model, as it relies on a data augmentation strategy that spawns objects with random positions and orientations. In contrast, our model achieves higher success rates due to the inherent robustness provided by the equivariant architecture.

\subsection{Impact of the Physics-Informed Refinement Layer}
To quantify the contribution of the physics-informed refinement layer, we compare the performance of our full model with a version lacking this component. From the results in Figure~\ref{fig:results}, we observe that the refinement layer significantly enhances grasp stability and diversity. This improvement is due to the layer's ability to guide the grasping process with gradients calculated from a physics simulator, ensuring the generated grasps are not only physically plausible but also $SE(3)$ equivariant. Additionally, as shown in Table~\ref{table:ablation}, the refinement layer can act as a plug-in for any iterative generative model, boosting performance by 5\%-10\% in terms of grasp success rate.
\begin{table}[t]
\centering
\begin{tabular}{ccc}
\toprule
               & \textbf{Ours} & \textbf{SceneDiffuser}~\cite{huang2023diffusion} \\ \midrule
\textbf{without Refinement}     & 61.42 & 60.47 \\ 
\textbf{with Refinement}  & 67.89 & 65.31 \\ \bottomrule
\end{tabular}
\captionsetup{belowskip=-20pt}
\caption{Comparison of the grasp success rate with and without physics-informed refinement layer for our model and SceneDiffuser.}
\label{table:ablation}
\end{table}

\subsection{Qualitative Analysis}
We present qualitative results to further assess the effectiveness of our method in generating stable grasps for unseen objects. As shown in Figure~\ref{fig:qual}, our method successfully produces stable and diverse grasps for objects not encountered during training. The figure also demonstrates the impact of the physics-informed refinement weight $\lambda$ on the type of grasp generated. With a stronger $\lambda$, our model tends to generate more stable power grasps suitable for holding objects securely. On the other hand, with a smaller $\lambda$, the model favors precision fingertip grasps, which are more delicate and precise. This flexibility highlights the model's ability to adapt to different grasping scenarios by tuning the refinement weight $\lambda$ to control how much the refinement can affect the reverse diffusion process.

\section{Discussion}
\label{sec:conclusion}
In this work, we introduce a symmetry-aware, diffusion-based framework for dexterous grasp generation leveraging geometric algebra. Our approach significantly improves generalization to data with OOD poses, enhances data and parameter efficiency, and produces physically plausible grasps through a physics-informed refinement layer. The model's robustness and flexibility make it well-suited for real-world robotic manipulation tasks, demonstrating substantial improvements over existing methods.

\textbf{Limitation and Future Works} While our GA-based architecture improves generalization and data efficiency, it imposes a notable training overhead. On an NVIDIA RTX A6000 GPU, our model takes roughly 7 days to train and consumes 2–3$\times$ more GPU memory than comparable non-equivariant diffusion-based methods. Additionally, the physics-refinement layer depends on watertight mesh reconstructions, which may fail with noisy normals to produce invalid contact points. We have also observed failure cases when the object’s size is significantly smaller than what was encountered during training.

Despite these constraints, the core architecture could be adapted to other tasks where $SE(3)$ symmetries are crucial. Future work could investigate more efficient GA representations to alleviate the current computational demands while broadening the scope of possible manipulation tasks.







\renewcommand*{\bibfont}{\small}
\bibliographystyle{IEEEtranN}
\bibliography{IEEEabrv,main}

\end{document}